\pdfoutput=1
\documentclass[sigconf]{acmart}

\usepackage{booktabs} % For formal tables
\usepackage{comment}
\usepackage{bm}
\usepackage{amsthm,amsmath,amsfonts,amssymb,dsfont}
\usepackage{balance}

\setcopyright{rightsretained}
\copyrightyear{2018}
\acmYear{2018}
\acmConference[Woodstock '18]{Woodstock '18: ACM Symposium on Neural Gaze Detection}{June 03--05, 2018}{Woodstock, NY}
\acmBooktitle{Woodstock '18: ACM Symposium on Neural Gaze Detection, June 03--05, 2018, Woodstock, NY}
\acmDOI{10.1145/1122445.1122456}
\acmISBN{978-1-4503-9999-9/18/06}

\acmArticle{4}
\acmPrice{15.00}

\begin{document}
\emergencystretch 1em

%\title{Neural Graph Learning on Image Representation with Instance-Level Semantic}
\title{Graph-RISE: Graph-Regularized Image Semantic Embedding}

\author{Da-Cheng Juan, Chun-Ta Lu, Zhen Li, Futang Peng, Aleksei Timofeev, Yi-Ting Chen, Yaxi Gao, Tom Duerig, Andrew Tomkins, Sujith Ravi\vspace{6pt}}

\affiliation{%
  \institution{Google AI}
  \city{Mountain View}
  \state{CA}
}
\email{{dacheng,chunta,zhenli,futangpeng,altimofeev,yitingchen,gyxlucy,tduerig,tomkins,sravi}@google.com}

\renewcommand{\shortauthors}{D.-C. Juan et al.}

\begin{abstract}
Learning image representations to capture fine-grained semantics has been a challenging and important task enabling many applications such as image search and clustering. In this paper, we present Graph-Regularized Image Semantic Embedding (Graph-RISE), a large-scale neural graph learning framework that allows us to train embeddings to discriminate an unprecedented $\mathcal{O}(40M)$ ultra-fine-grained semantic labels.  Graph-RISE outperforms state-of-the-art image embedding algorithms on several evaluation tasks, including image classification and triplet ranking. We provide case studies to demonstrate that, qualitatively, image retrieval based on Graph-RISE effectively captures semantics and, compared to the state-of-the-art, differentiates nuances at levels that are closer to human-perception.
\end{abstract}






\maketitle

%%%%% MISC %%%%%

\newcommand\reminder[1]{\textcolor{blue}{#1}}
\newcommand{\ie}{{\textit i.e.}}
\newcommand{\eg}{{\textit e.g.}}
\newcommand{\coclickgraph}{co-graph~}
\newcommand{\vsigraph}{vis-graph~}

\newcommand{\ourmodel}[0]{Graph-RISE}

%%%%% NEW MATH DEFINITIONS %%%%%

% Mark sections of captions for referring to divisions of figures
\newcommand{\figleft}{{\em (Left)}}
\newcommand{\figcenter}{{\em (Center)}}
\newcommand{\figright}{{\em (Right)}}
\newcommand{\figtop}{{\em (Top)}}
\newcommand{\figbottom}{{\em (Bottom)}}
\newcommand{\captiona}{{\em (a)}}
\newcommand{\captionb}{{\em (b)}}
\newcommand{\captionc}{{\em (c)}}
\newcommand{\captiond}{{\em (d)}}

% Highlight a newly defined term
\newcommand{\newterm}[1]{{\bf #1}}

% Figure reference, lower-case.
\def\figref#1{figure~\ref{#1}}
% Figure reference, capital. For start of sentence
\def\Figref#1{Figure~\ref{#1}}
\def\twofigref#1#2{figures \ref{#1} and \ref{#2}}
\def\quadfigref#1#2#3#4{figures \ref{#1}, \ref{#2}, \ref{#3} and \ref{#4}}
% Section reference, lower-case.
\def\secref#1{section~\ref{#1}}
% Section reference, capital.
\def\Secref#1{Section~\ref{#1}}
% Reference to two sections.
\def\twosecrefs#1#2{sections \ref{#1} and \ref{#2}}
% Reference to three sections.
\def\secrefs#1#2#3{sections \ref{#1}, \ref{#2} and \ref{#3}}
% Reference to an equation, lower-case.
\def\eqref#1{equation~\ref{#1}}
% Reference to an equation, upper case
\def\Eqref#1{Equation~\ref{#1}}
% A raw reference to an equation---avoid using if possible
\def\plaineqref#1{\ref{#1}}
% Reference to a chapter, lower-case.
\def\chapref#1{chapter~\ref{#1}}
% Reference to an equation, upper case.
\def\Chapref#1{Chapter~\ref{#1}}
% Reference to a range of chapters
\def\rangechapref#1#2{chapters\ref{#1}--\ref{#2}}
% Reference to an algorithm, lower-case.
\def\algref#1{algorithm~\ref{#1}}
% Reference to an algorithm, upper case.
\def\Algref#1{Algorithm~\ref{#1}}
\def\twoalgref#1#2{algorithms \ref{#1} and \ref{#2}}
\def\Twoalgref#1#2{Algorithms \ref{#1} and \ref{#2}}
% Reference to a part, lower case
\def\partref#1{part~\ref{#1}}
% Reference to a part, upper case
\def\Partref#1{Part~\ref{#1}}
\def\twopartref#1#2{parts \ref{#1} and \ref{#2}}

\def\ceil#1{\lceil #1 \rceil}
\def\floor#1{\lfloor #1 \rfloor}
\def\1{\bm{1}}
\newcommand{\train}{\mathcal{D}}
\newcommand{\valid}{\mathcal{D_{\mathrm{valid}}}}
\newcommand{\test}{\mathcal{D_{\mathrm{test}}}}

\def\eps{{\epsilon}}

% Random variables
\def\reta{{\textnormal{$\eta$}}}
\def\ra{{\textnormal{a}}}
\def\rb{{\textnormal{b}}}
\def\rc{{\textnormal{c}}}
\def\rd{{\textnormal{d}}}
\def\re{{\textnormal{e}}}
\def\rf{{\textnormal{f}}}
\def\rg{{\textnormal{g}}}
\def\rh{{\textnormal{h}}}
\def\ri{{\textnormal{i}}}
\def\rj{{\textnormal{j}}}
\def\rk{{\textnormal{k}}}
\def\rl{{\textnormal{l}}}
% rm is already a command, just don't name any random variables m
\def\rn{{\textnormal{n}}}
\def\ro{{\textnormal{o}}}
\def\rp{{\textnormal{p}}}
\def\rq{{\textnormal{q}}}
\def\rr{{\textnormal{r}}}
\def\rs{{\textnormal{s}}}
\def\rt{{\textnormal{t}}}
\def\ru{{\textnormal{u}}}
\def\rv{{\textnormal{v}}}
\def\rw{{\textnormal{w}}}
\def\rx{{\textnormal{x}}}
\def\ry{{\textnormal{y}}}
\def\rz{{\textnormal{z}}}

% Random vectors
\def\rvepsilon{{\mathbf{\epsilon}}}
\def\rvtheta{{\mathbf{\theta}}}
\def\rva{{\mathbf{a}}}
\def\rvb{{\mathbf{b}}}
\def\rvc{{\mathbf{c}}}
\def\rvd{{\mathbf{d}}}
\def\rve{{\mathbf{e}}}
\def\rvf{{\mathbf{f}}}
\def\rvg{{\mathbf{g}}}
\def\rvh{{\mathbf{h}}}
\def\rvu{{\mathbf{i}}}
\def\rvj{{\mathbf{j}}}
\def\rvk{{\mathbf{k}}}
\def\rvl{{\mathbf{l}}}
\def\rvm{{\mathbf{m}}}
\def\rvn{{\mathbf{n}}}
\def\rvo{{\mathbf{o}}}
\def\rvp{{\mathbf{p}}}
\def\rvq{{\mathbf{q}}}
\def\rvr{{\mathbf{r}}}
\def\rvs{{\mathbf{s}}}
\def\rvt{{\mathbf{t}}}
\def\rvu{{\mathbf{u}}}
\def\rvv{{\mathbf{v}}}
\def\rvw{{\mathbf{w}}}
\def\rvx{{\mathbf{x}}}
\def\rvy{{\mathbf{y}}}
\def\rvz{{\mathbf{z}}}

% Elements of random vectors
\def\erva{{\textnormal{a}}}
\def\ervb{{\textnormal{b}}}
\def\ervc{{\textnormal{c}}}
\def\ervd{{\textnormal{d}}}
\def\erve{{\textnormal{e}}}
\def\ervf{{\textnormal{f}}}
\def\ervg{{\textnormal{g}}}
\def\ervh{{\textnormal{h}}}
\def\ervi{{\textnormal{i}}}
\def\ervj{{\textnormal{j}}}
\def\ervk{{\textnormal{k}}}
\def\ervl{{\textnormal{l}}}
\def\ervm{{\textnormal{m}}}
\def\ervn{{\textnormal{n}}}
\def\ervo{{\textnormal{o}}}
\def\ervp{{\textnormal{p}}}
\def\ervq{{\textnormal{q}}}
\def\ervr{{\textnormal{r}}}
\def\ervs{{\textnormal{s}}}
\def\ervt{{\textnormal{t}}}
\def\ervu{{\textnormal{u}}}
\def\ervv{{\textnormal{v}}}
\def\ervw{{\textnormal{w}}}
\def\ervx{{\textnormal{x}}}
\def\ervy{{\textnormal{y}}}
\def\ervz{{\textnormal{z}}}

% Random matrices
\def\rmA{{\mathbf{A}}}
\def\rmB{{\mathbf{B}}}
\def\rmC{{\mathbf{C}}}
\def\rmD{{\mathbf{D}}}
\def\rmE{{\mathbf{E}}}
\def\rmF{{\mathbf{F}}}
\def\rmG{{\mathbf{G}}}
\def\rmH{{\mathbf{H}}}
\def\rmI{{\mathbf{I}}}
\def\rmJ{{\mathbf{J}}}
\def\rmK{{\mathbf{K}}}
\def\rmL{{\mathbf{L}}}
\def\rmM{{\mathbf{M}}}
\def\rmN{{\mathbf{N}}}
\def\rmO{{\mathbf{O}}}
\def\rmP{{\mathbf{P}}}
\def\rmQ{{\mathbf{Q}}}
\def\rmR{{\mathbf{R}}}
\def\rmS{{\mathbf{S}}}
\def\rmT{{\mathbf{T}}}
\def\rmU{{\mathbf{U}}}
\def\rmV{{\mathbf{V}}}
\def\rmW{{\mathbf{W}}}
\def\rmX{{\mathbf{X}}}
\def\rmY{{\mathbf{Y}}}
\def\rmZ{{\mathbf{Z}}}

% Elements of random matrices
\def\ermA{{\textnormal{A}}}
\def\ermB{{\textnormal{B}}}
\def\ermC{{\textnormal{C}}}
\def\ermD{{\textnormal{D}}}
\def\ermE{{\textnormal{E}}}
\def\ermF{{\textnormal{F}}}
\def\ermG{{\textnormal{G}}}
\def\ermH{{\textnormal{H}}}
\def\ermI{{\textnormal{I}}}
\def\ermJ{{\textnormal{J}}}
\def\ermK{{\textnormal{K}}}
\def\ermL{{\textnormal{L}}}
\def\ermM{{\textnormal{M}}}
\def\ermN{{\textnormal{N}}}
\def\ermO{{\textnormal{O}}}
\def\ermP{{\textnormal{P}}}
\def\ermQ{{\textnormal{Q}}}
\def\ermR{{\textnormal{R}}}
\def\ermS{{\textnormal{S}}}
\def\ermT{{\textnormal{T}}}
\def\ermU{{\textnormal{U}}}
\def\ermV{{\textnormal{V}}}
\def\ermW{{\textnormal{W}}}
\def\ermX{{\textnormal{X}}}
\def\ermY{{\textnormal{Y}}}
\def\ermZ{{\textnormal{Z}}}

% Vectors
\def\vzero{{\bm{0}}}
\def\vone{{\bm{1}}}
\def\vmu{{\bm{\mu}}}
\def\vtheta{{\bm{\theta}}}
\def\va{{\bm{a}}}
\def\vb{{\bm{b}}}
\def\vc{{\bm{c}}}
\def\vd{{\bm{d}}}
\def\ve{{\bm{e}}}
\def\vf{{\bm{f}}}
\def\vg{{\bm{g}}}
\def\vh{{\bm{h}}}
\def\vi{{\bm{i}}}
\def\vj{{\bm{j}}}
\def\vk{{\bm{k}}}
\def\vl{{\bm{l}}}
\def\vm{{\bm{m}}}
\def\vn{{\bm{n}}}
\def\vo{{\bm{o}}}
\def\vp{{\bm{p}}}
\def\vq{{\bm{q}}}
\def\vr{{\bm{r}}}
\def\vs{{\bm{s}}}
\def\vt{{\bm{t}}}
\def\vu{{\bm{u}}}
\def\vv{{\bm{v}}}
\def\vw{{\bm{w}}}
\def\vx{{\bm{x}}}
\def\vy{{\bm{y}}}
\def\vz{{\bm{z}}}
\def\vtheta{{\bm{\theta}}}

% Elements of vectors
\def\evalpha{{\alpha}}
\def\evbeta{{\beta}}
\def\evepsilon{{\epsilon}}
\def\evlambda{{\lambda}}
\def\evomega{{\omega}}
\def\evmu{{\mu}}
\def\evpsi{{\psi}}
\def\evsigma{{\sigma}}
\def\evtheta{{\theta}}
\def\eva{{a}}
\def\evb{{b}}
\def\evc{{c}}
\def\evd{{d}}
\def\eve{{e}}
\def\evf{{f}}
\def\evg{{g}}
\def\evh{{h}}
\def\evi{{i}}
\def\evj{{j}}
\def\evk{{k}}
\def\evl{{l}}
\def\evm{{m}}
\def\evn{{n}}
\def\evo{{o}}
\def\evp{{p}}
\def\evq{{q}}
\def\evr{{r}}
\def\evs{{s}}
\def\evt{{t}}
\def\evu{{u}}
\def\evv{{v}}
\def\evw{{w}}
\def\evx{{x}}
\def\evy{{y}}
\def\evz{{z}}

% Matrix
\def\mA{{\bm{A}}}
\def\mB{{\bm{B}}}
\def\mC{{\bm{C}}}
\def\mD{{\bm{D}}}
\def\mE{{\bm{E}}}
\def\mF{{\bm{F}}}
\def\mG{{\bm{G}}}
\def\mH{{\bm{H}}}
\def\mI{{\bm{I}}}
\def\mJ{{\bm{J}}}
\def\mK{{\bm{K}}}
\def\mL{{\bm{L}}}
\def\mM{{\bm{M}}}
\def\mN{{\bm{N}}}
\def\mO{{\bm{O}}}
\def\mP{{\bm{P}}}
\def\mQ{{\bm{Q}}}
\def\mR{{\bm{R}}}
\def\mS{{\bm{S}}}
\def\mT{{\bm{T}}}
\def\mU{{\bm{U}}}
\def\mV{{\bm{V}}}
\def\mW{{\bm{W}}}
\def\mX{{\bm{X}}}
\def\mY{{\bm{Y}}}
\def\mZ{{\bm{Z}}}
\def\mBeta{{\bm{\beta}}}
\def\mPhi{{\bm{\Phi}}}
\def\mLambda{{\bm{\Lambda}}}
\def\mSigma{{\bm{\Sigma}}}

% Tensor
\newcommand{\tens}[1]{\bm{\mathsfit{#1}}}
\def\tA{{\tens{A}}}
\def\tB{{\tens{B}}}
\def\tC{{\tens{C}}}
\def\tD{{\tens{D}}}
\def\tE{{\tens{E}}}
\def\tF{{\tens{F}}}
\def\tG{{\tens{G}}}
\def\tH{{\tens{H}}}
\def\tI{{\tens{I}}}
\def\tJ{{\tens{J}}}
\def\tK{{\tens{K}}}
\def\tL{{\tens{L}}}
\def\tM{{\tens{M}}}
\def\tN{{\tens{N}}}
\def\tO{{\tens{O}}}
\def\tP{{\tens{P}}}
\def\tQ{{\tens{Q}}}
\def\tR{{\tens{R}}}
\def\tS{{\tens{S}}}
\def\tT{{\tens{T}}}
\def\tU{{\tens{U}}}
\def\tV{{\tens{V}}}
\def\tW{{\tens{W}}}
\def\tX{{\tens{X}}}
\def\tY{{\tens{Y}}}
\def\tZ{{\tens{Z}}}

% Graph
\def\gA{{\mathcal{A}}}
\def\gB{{\mathcal{B}}}
\def\gC{{\mathcal{C}}}
\def\gD{{\mathcal{D}}}
\def\gE{{\mathcal{E}}}
\def\gF{{\mathcal{F}}}
\def\gG{{\mathcal{G}}}
\def\gH{{\mathcal{H}}}
\def\gI{{\mathcal{I}}}
\def\gJ{{\mathcal{J}}}
\def\gK{{\mathcal{K}}}
\def\gL{{\mathcal{L}}}
\def\gM{{\mathcal{M}}}
\def\gN{{\mathcal{N}}}
\def\gO{{\mathcal{O}}}
\def\gP{{\mathcal{P}}}
\def\gQ{{\mathcal{Q}}}
\def\gR{{\mathcal{R}}}
\def\gS{{\mathcal{S}}}
\def\gT{{\mathcal{T}}}
\def\gU{{\mathcal{U}}}
\def\gV{{\mathcal{V}}}
\def\gW{{\mathcal{W}}}
\def\gX{{\mathcal{X}}}
\def\gY{{\mathcal{Y}}}
\def\gZ{{\mathcal{Z}}}

% Sets
\def\sA{{\mathbb{A}}}
\def\sB{{\mathbb{B}}}
\def\sC{{\mathbb{C}}}
\def\sD{{\mathbb{D}}}
% Don't use a set called E, because this would be the same as our symbol
% for expectation.
\def\sF{{\mathbb{F}}}
\def\sG{{\mathbb{G}}}
\def\sH{{\mathbb{H}}}
\def\sI{{\mathbb{I}}}
\def\sJ{{\mathbb{J}}}
\def\sK{{\mathbb{K}}}
\def\sL{{\mathbb{L}}}
\def\sM{{\mathbb{M}}}
\def\sN{{\mathbb{N}}}
\def\sO{{\mathbb{O}}}
\def\sP{{\mathbb{P}}}
\def\sQ{{\mathbb{Q}}}
\def\sR{{\mathbb{R}}}
\def\sS{{\mathbb{S}}}
\def\sT{{\mathbb{T}}}
\def\sU{{\mathbb{U}}}
\def\sV{{\mathbb{V}}}
\def\sW{{\mathbb{W}}}
\def\sX{{\mathbb{X}}}
\def\sY{{\mathbb{Y}}}
\def\sZ{{\mathbb{Z}}}

% Entries of a matrix
\def\emLambda{{\Lambda}}
\def\emA{{A}}
\def\emB{{B}}
\def\emC{{C}}
\def\emD{{D}}
\def\emE{{E}}
\def\emF{{F}}
\def\emG{{G}}
\def\emH{{H}}
\def\emI{{I}}
\def\emJ{{J}}
\def\emK{{K}}
\def\emL{{L}}
\def\emM{{M}}
\def\emN{{N}}
\def\emO{{O}}
\def\emP{{P}}
\def\emQ{{Q}}
\def\emR{{R}}
\def\emS{{S}}
\def\emT{{T}}
\def\emU{{U}}
\def\emV{{V}}
\def\emW{{W}}
\def\emX{{X}}
\def\emY{{Y}}
\def\emZ{{Z}}
\def\emSigma{{\Sigma}}

% entries of a tensor
% Same font as tensor, without \bm wrapper
\newcommand{\etens}[1]{\mathsfit{#1}}
\def\etLambda{{\etens{\Lambda}}}
\def\etA{{\etens{A}}}
\def\etB{{\etens{B}}}
\def\etC{{\etens{C}}}
\def\etD{{\etens{D}}}
\def\etE{{\etens{E}}}
\def\etF{{\etens{F}}}
\def\etG{{\etens{G}}}
\def\etH{{\etens{H}}}
\def\etI{{\etens{I}}}
\def\etJ{{\etens{J}}}
\def\etK{{\etens{K}}}
\def\etL{{\etens{L}}}
\def\etM{{\etens{M}}}
\def\etN{{\etens{N}}}
\def\etO{{\etens{O}}}
\def\etP{{\etens{P}}}
\def\etQ{{\etens{Q}}}
\def\etR{{\etens{R}}}
\def\etS{{\etens{S}}}
\def\etT{{\etens{T}}}
\def\etU{{\etens{U}}}
\def\etV{{\etens{V}}}
\def\etW{{\etens{W}}}
\def\etX{{\etens{X}}}
\def\etY{{\etens{Y}}}
\def\etZ{{\etens{Z}}}

% The true underlying data generating distribution
\newcommand{\pdata}{p_{\rm{data}}}
% The empirical distribution defined by the training set
\newcommand{\ptrain}{\hat{p}_{\rm{data}}}
\newcommand{\Ptrain}{\hat{P}_{\rm{data}}}
% The model distribution
\newcommand{\pmodel}{p_{\rm{model}}}
\newcommand{\Pmodel}{P_{\rm{model}}}
\newcommand{\ptildemodel}{\tilde{p}_{\rm{model}}}
% Stochastic autoencoder distributions
\newcommand{\pencode}{p_{\rm{encoder}}}
\newcommand{\pdecode}{p_{\rm{decoder}}}
\newcommand{\precons}{p_{\rm{reconstruct}}}

\newcommand{\laplace}{\mathrm{Laplace}} % Laplace distribution

\newcommand{\E}{\mathbb{E}}
\newcommand{\Ls}{\mathcal{L}}
\newcommand{\R}{\mathbb{R}}
\newcommand{\emp}{\tilde{p}}
\newcommand{\lr}{\alpha}
\newcommand{\reg}{\lambda}
\newcommand{\rect}{\mathrm{rectifier}}
\newcommand{\softmax}{\mathrm{softmax}}
\newcommand{\sigmoid}{\sigma}
\newcommand{\softplus}{\zeta}
\newcommand{\KL}{D_{\mathrm{KL}}}
\newcommand{\Var}{\mathrm{Var}}
\newcommand{\standarderror}{\mathrm{SE}}
\newcommand{\Cov}{\mathrm{Cov}}
% Wolfram Mathworld says $L^2$ is for function spaces and $\ell^2$ is for vectors
% But then they seem to use $L^2$ for vectors throughout the site, and so does
% wikipedia.
\newcommand{\normlzero}{L^0}
\newcommand{\normlone}{L^1}
\newcommand{\normltwo}{L^2}
\newcommand{\normlp}{L^p}
\newcommand{\normmax}{L^\infty}

\newcommand{\parents}{Pa} % See usage in notation.tex. Chosen to match Daphne's book.
\let\ab\allowbreak

\section{Introduction}
\label{sec:introduction}
\begin{comment}
\textit{``How to learn image embeddings that capture fine-grained semantics based on the instance of an image?'' ``Is it possible for such embeddings to further understand image semantics closer to humans' perception?''}  Learning image embeddings that capture fine-grained semantics is the core of many modern image-related applications such as image search, either querying by traditional keywords or by a query image~\cite{murphy2015visual}. However, learning such embeddings is a challenging task, partly due to the large variations among images that belong to the same category or class. For example, within the same semantic class, many factors such as color, shape or texture can still be scaled or translated within an image.
\end{comment}

\textit{``Is it possible to learn image content descriptors (a.k.a., embeddings) that capture image semantics and similarity close to human perception?''} Learning image embeddings that capture fine-grained semantics is the core of many modern image-related applications such as image search, either querying by traditional keywords or by an example query image \cite{murphy2015visual}. Learning such embeddings is a challenging task, partly due to the large variations seen among images that belong to the same category or class.

\begin{figure}[tbp]
  \includegraphics[scale=0.8,bb=0 0 304 169]{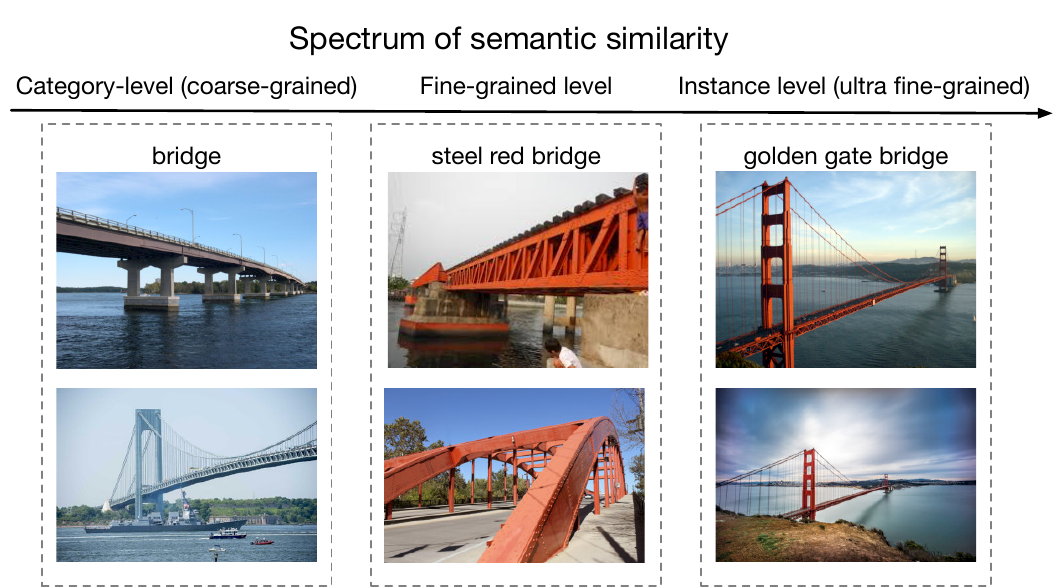}
  \caption{Spectrum of image semantic similarity. We provide six image examples (two for each granularity) to illustrate the difference from coarser (left) to ultra-fine granularity (right). We refer to ultra fine-grained as ``instance-level'' to contrast with category-level and fine-grained semantics.}
  \label{fig:semantic_granularity}
\end{figure}

Several previous works consider category-level image semantics \cite{hadsell2006dimensionality,taylor2011learning}, in which two images are considered semantically similar if they belong to the same category. As illustrated in Figure~\ref{fig:semantic_granularity}, category-level similarity may not be sufficient for modern vision-based applications such as query-by-image, which often require the distinction of nuances among images within the same category.

Recently, deep ranking models \cite{wang2014learning,wu2017sampling} have been proposed to learn fine-grained image similarity. These ranking models are trained with image triplets, where each entry contains \{query image, positive image, negative image\}. The goal is to rank (query, positive image) as more similar than (query, negative image); this training formulation can encode distinctions that are as fine-grained as the construction of the triplets allows. In practice, however, it becomes increasingly difficult to generate a large corpus of triplets that encode sufficiently fine-grained distinctions by ensuring the negative image is similar enough to the query image, but not too similar. Furthermore, since human raters need to be involved to provide the triplet ranking ground truth, collecting high quality image triplets for training is costly and labor-intensive.  We instead propose moving from triplet learning to a classification framework that learns embeddings capable of associating an image to one of a large number of possible query strings.

Such an approach produces image embeddings that are predictive of queries that might lead to the image.  In addition, we also obtain similarity data between the images themselves, encoding for example the fact that two images were both clicked in a particular setting.  This relational data encodes important aspects of human image perception but is not easily encapsulated by labels (\ie, image-label pairs for training).  To incorporate image-image similarity into the training we employ neural graph learning, in which the model is trained by minimizing the supervised loss combined with a graph regularizer that drives the model to reduce embedding distance between similar image pairs.

To the best of our knowledge, this work brings the following contributions:
\begin{itemize}
    \item{{\bf Effective embedding learning via large scale classification.} We formulate the problem of image embedding learning as an image classification task at an unprecedented scale, with label space (\ie, total number of classes) in $\gO (40M)$ and the number of images in $\gO (260M)$. This is {\bf the largest scale} in terms of number of classes, and one of the largest in terms of images used for learning image embeddings. Furthermore, the proposed model is one of the {\bf largest vision models} in terms of the number of parameters (see Section~\ref{subsec:Network Architecture} and Section~\ref{subsec:Training Infrastructure}). No previous literature has demonstrated the effectiveness of such a large-scale image classification for learning image representation.}
    \item{{\bf Neural graph learning on image representation.} We propose a neural graph learning framework that leverages graph structure to regularize the training of deep neural networks. This is the first work deploying large-scale neural graph learning for image representation. We will describe below two techniques to construct image-image graphs based on ``co-click'' rate and ``similar-image click'' rate, designed to capture ultra-fine-grained notions of similarity that emerge from human perception of result sets.} 
     %We also propose two methods---based on ``co-click'' rate and ``similar-image click'' rate---to construct graphs that capture image-image similarity closer to human perception.
    \item{{\bf {\ourmodel} for instance-level semantics.} We present {\ourmodel}, an image embedding that captures ultra-fine-grained, instance-level semantics. {\ourmodel} outperforms the state-of-the-art algorithms for learning image embeddings on several evaluations based on k-Nearest-Neighbor (kNN) search and triplet ranking. Experimental results show that {\ourmodel} improves the Top-1 accuracy of the kNN evaluation by approximately 2X on the ImageNet dataset and by more than 5X on the iNaturalist dataset. Case studies also show that, qualitatively, {\ourmodel} outperforms the state of the art and captures instance-level semantics. }
\end{itemize}

The remainder of this paper is organized as follows. Section~\ref{sec:related_work} provides related work on learning image embeddings. Section~\ref{sec:Problem Formulation} formulates the problem and provides the details of training datasets. Section~\ref{sec:overview} explains the proposed learning algorithms, followed by Section~\ref{sec:architecture} with the details of network architecture and training infrastructure. Section~\ref{sec:experiment} shows the experimental results and Section~\ref{sec:conclusion} concludes this paper.

\section{Related Work}
\label{sec:related_work}
There are several prior works on learning image similarity~\cite{wang2009learning,guillaumin2009tagprop,taylor2011learning}. Most of them focus on category-level image similarity, 
in which two images are considered to be similar if they belong to the same category. In general, visual and semantic similarities tend to be consistent with each other across category boundaries~\cite{deselaers2011visual}.  Visual variability within a semantically-defined category still exists, especially for broadly defined categories such as ``animal,'' or ``plant,''  as a result of the broad semantic distinctions within such classes.  As classes become finer grained, however, the visual distinctions within a class due to natural variations in image capture (angle, lighting, background, etc) become larger relative to the fine distinctions between  classes that are semantically closer; hence, new techniques are required.

For learning fine-grained image similarity, local distance learning \cite{frome2007image} and OASIS \cite{chechik2010large} developed ranking models based on hand-crafted features, trained with triplets wherein each entry contains \{query image, positive image, negative image\} that characterizes the ranking orders based on relative similarity. In \cite{wang2014learning}, a DeepRanking model that integrates the deep learning and ranking model is proposed to learn a fine-grained image similarity ranking model directly from images, rather than from hand-crafted features. As discussed above, while these ranking models have been widely used for learning image embeddings, the model performance relies heavily on the quality of triplet samples, which involves pair-wise comparisons that can be costly and labor-intensive to collect. As we will show later in Section~\ref{sec:Problem Formulation} and Section~\ref{sec:experiment}, {\ourmodel} does not require models to be trained by triplets and outperforms the state-of-the-art on capturing image semantics for several evaluation tasks. %Further, the case studies (in Section~\ref{subsec:Case Studies}) demonstrate that {\ourmodel} is able to capture instance-level, fine-grained semantics.

There has also been a significant amount of work on improving image classification to near-human levels ~\cite{russakovsky2015imagenet} by increasing the representational capacity and the depth of network architectures.  See, \eg, VGG-19~\cite{simonyan2014very}, Inception~\cite{szegedy2016rethinking}, and ResNet~\cite{he2016deep}. To support learning such deep networks with millions of parameters, large-scale datasets such as ImageNet~\cite{krizhevsky2012imagenet},  iNaturalist~\cite{van2018inaturalist} and YouTube-8M~\cite{abu2016youtube} have played a crucial role. For example, the authors of ~\cite{oquab2014learning} demonstrate that the rich mid-level image features learned by Convolutional Neural Networks (CNNs) on ImageNet can be efficiently transferred to other visual recognition tasks with limited amount of training data. Their study suggests that the number of images and the coverage of classes for training in the source task are important for the performance in the target task. In ~\cite{sun2017revisiting}, the authors reveal a logarithmic relationship between the performance on vision tasks and the amount of training data used for representation learning. In this paper, we share the same observation and further show that when increasing the number of classes to $\gO (40M)$ with sufficient amount of training data, the purposed {\ourmodel} is able to capture instance-level, ultra-fine-grained semantics.

\section{Problem Formulation}
\label{sec:Problem Formulation}
In this section, we formulate the task of learning an image embedding model, and then provide the details of training dataset used for this task.

\paragraph{Problem Formulation}
Given the following inputs: 
\begin{itemize}
    \item A labeled set $\gD_L$ that contains image-label pairs $(x, y)$, where label $y$ provides ultra-fine-grained semantics to the corresponding image $x$.
    \item An unlabeled set $\gD_U$ that contains images without labels.
    % \item A hypothesis space $\gH$ where each model is parameterized by $\vtheta$.
\end{itemize}
The objective is to find an image embedding model that achieves instance-level semantic understanding. Specifically, let $\phi(\cdot)$ represent a function that projects an image to a dense vector representing an embedding; given two images $x_1$ and $x_2$, the similarity in the image embedding space is defined according to a distance metric: $d(\phi(x_1), \phi(x_2))$, where $d(\cdot)$ is a distance function (\eg, Euclidean distance or cosine distance). If $x_1$ and $x_2$ belong to the same class, an ideal $\phi(\cdot)$ minimizes the distance between $\phi(x_1)$ and $\phi(x_2)$, indicating these two images to be semantically similar. Table~\ref{tbl:symbol_table} provides the symbols and the corresponding definitions used throughout this paper.

\begin{table}[tbp]
\small
\caption{List of symbols.}
\label{tbl:symbol_table}
\begin{tabular}{ll}
\toprule
Symbol & Definition and description\\
	\hline
	$(x, y)$	 & Image-label pair \\ 
	$\gD_L$, $\gD_U$	 & Labeled dataset, unlabeled dataset \\
	$\gK$ & Total number of classes  \\ 
    $\gE$ & Set of edges \\ 
    $w_{u,v}$ & Edge weight \\ 
    $\eta$ & Margin for triplet distance \\
    $\phi(\cdot)$ & Embedding mapping function  \\ 
    $d(\cdot)$ & Distance function  \\ 
    $p(\cdot), q(\cdot)$ & Probability and ground-truth distribution \\  
    $\gL(\cdot)$ & Loss function \\ 
    $\Omega(\cdot)$ & Graph regularizer \\     
    $\vtheta$ & Model parameters \\ 
    % $z_i$ & Logits or unnormalized log probability  \\ 
    % $\mW$, $b$ & Weights and bias for target label \\
\bottomrule    
\end{tabular}
\end{table}

% \begin{table}[tb]
% 	\centering
% 	\caption{Symbols and definitions}
% 	%\begin{tabular}{|c||l|}
% 	\begin{tabular}{|p{0.15\linewidth}||p{0.75\linewidth}|}	
% 	\hline 
% 	Symbol& Definition\\
% 	\hline\hline
% 	$(x, y)$	 & Image-label pair \\ \hline
% 	$\gD_L$, $\gD_U$	 & Labeled dataset, unlabeled dataset \\ \hline
%     $\phi(\cdot)$ & Embedding mapping function  \\ \hline	
%     $d(\cdot)$ & Distance function  \\ \hline
%     $K$ & Total number of classes  \\ \hline
%     $z_i$ & Logits or unnormalized log probability  \\ \hline
%     $\mW$, $b$ & Weights and bias for target label \\ \hline    
%     $p(\cdot), q(\cdot)$ & Probability and ground-truth distribution \\ \hline    
%     $\vtheta$ & Model parameters \\ \hline     
%     $\gL(\cdot)$ & Loss function \\ \hline      
%     $\gE$ & Set of edges \\ \hline  
%     $w_{u,v}$ & Edge weight \\ \hline  
%     $\Omega(\cdot)$ & Graph regularizer \\ \hline  
% 	\end{tabular}	
% 	\label{tbl:symbol_table}
% \end{table}

\paragraph{Dataset}
% \reminder{Change lamborghini to other images due to PR.}
\begin{comment}
The JFT-300M dataset is closely related and derived from the data which powers the Google Images. In this version, the dataset has 300M images and 375M labels, on average each image has 1.26 labels. These images are labeled with 18291 categories: e.g., 1165 type of animals and 5720 types of vehicles are labeled in the dataset. These categories form a rich hierarchy with the maximum depth of hierarchy being 12 and maximum number of child for parent node being 2876. The images are labeled using an algorithm that uses complex mixture of raw web signals, connections between webpages and user feedback. The algorithm starts from over one billion image label pairs, and ends up with 375M labels for 300M images with the aim to select labeled images with high precision.
\end{comment}

\begin{figure}[tbp]
  \includegraphics[scale=0.5,bb=0 0 372 406]{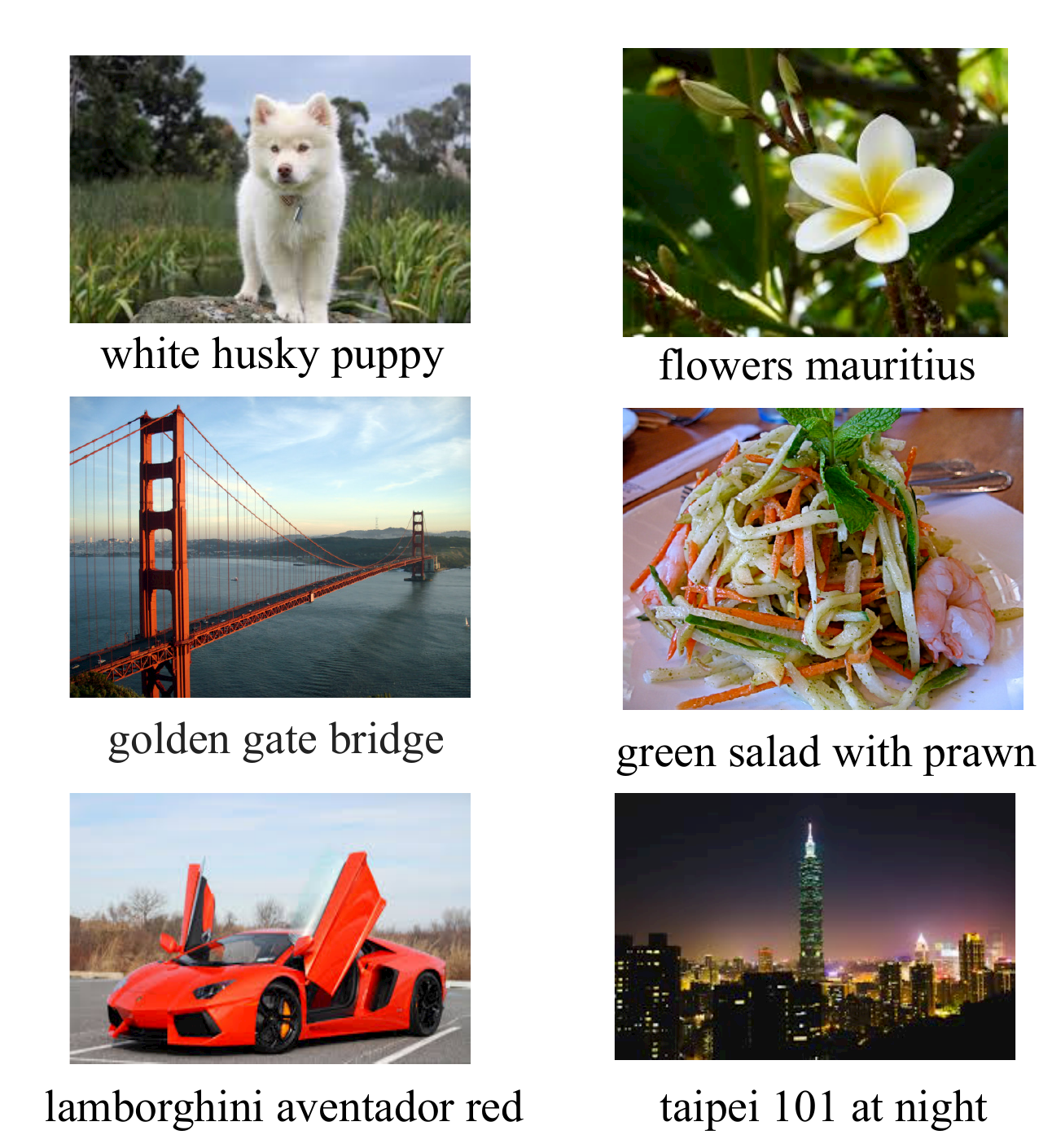}
  \caption{Six potential samples of image-query pairs. Each image is labeled with the corresponding textual search query.}
  \label{fig:six_training_samples}
\end{figure}

In order to achieve instance-level semantic understanding, the classes should be ultra-fine-grained. Thus, we created a training dataset $\gD_L$ derived from Google Image Search. It contains approximately 260 million images; selection data are used to characterize how frequently anonymous users selected an image when that image appeared in the results of a particular textual search query. After the characterization, a search query is then treated as the ``label'' annotated to the corresponding image to provide semantics as labeled samples. Each image is labeled with one or more queries (2.55 queries per image on average), and the total number of unique queries (used as classes) is around forty million. Figure~\ref{fig:six_training_samples} illustrates six potential image-query pairs\footnote{In this paper, ``image-query pairs'' and ``image-label pairs'' are used interchangeably since queries are used as labels.}. To the best of our knowledge, this is the largest scale of training data for learning image embedding in terms of the number of classes\footnote{In prior arts (such as \cite{wang2014learning}), the training dataset contains up to $\gO (15M)$ samples with $\gO (100K)$ classes.}, and one of the largest in terms of the number of training images \cite{sun2017revisiting,wang2014learning,schroff2015facenet}.

%\reminder{Give some statistics of the image-query pairs.}

Unlabeled dataset $\gD_U$ contains approximately 20 million images, without any annotation or labeling. Unlabeled dataset is mainly for constructing similarity graphs (see Section~\ref{subsec:Neural Graph Learning}) and for evaluation purposes (see Section~\ref{subsec:Evaluation Setup}).

\begin{comment}
The training system 300 can jointly train the image embedding model 100 and the text embedding model 200 to generate similar embeddings of the search query and the image if the selection data indicates that users frequently select the image when it is identified by a search result for the search query. 
A similarity between embeddings can be determined using any appropriate numerical similarity measure (e.g., a Euclidean distance, if the image embedding model 100 and the text embedding model 200 are configured to generate embeddings of the same dimensionality).
How should we share the data? Especially Navboost \& Coclick signals. (find some references that are publicly available?) (Tom said we should not use the term "click", "coclick")
\end{comment}

\section{Learning Algorithms}
\label{sec:overview}
In this section, we first introduce the algorithm that takes image-query pairs as training data in a supervised learning manner. Then we elaborate on neural graph learning, a methodology incorporating graph signals into the learning algorithm.

\subsection{Discriminative Embedding Learning}
\label{subsec:Supervised Learning}
\begin{comment}
\begin{itemize}
    \item{softmax classification loss}
    \item{sampled softmax classification loss}
    \item{sampled softmax classification loss + label smoothing}
    \item{classification loss + graph loss}
\end{itemize}
\end{comment}

\paragraph{From triplet loss to softmax loss.} In order to train image embedding models, metric learning objectives (such as contrastive loss~\cite{hadsell2006dimensionality,sun2014deep} and triplet loss~\cite{wang2014learning}), and classification objectives (such as logistic loss and softmax loss) have been widely explored. When using metric learning objectives, collecting high quality samples (\eg, triplets) is often challenging. Furthermore, optimizing metric learning objectives suffers from slow convergence or poor local optima if sampling techniques are inappropriately applied~\cite{wu2017sampling,sohn2016improved}. In order to achieve instance-level semantic understanding, we employ softmax loss for training the image embedding model. When the size of classes is sufficiently large, $\gO(40M)$ in our case, classification training (with softmax loss) works better than triplet loss \cite{wang2014learning}.  

For each training example $x$, the probability of each label $k \in \{1, \dots, \gK\}$ in our model is computed via softmax: 
\begin{equation}
p(k | x) = \frac{exp\{z_k\}}{\sum_{i = 1}^{\gK} exp\{z_i\}}
\end{equation}
where $z_i$ are the logits or unnormalized log probabilities. Here, the $z_i$ are computed by adding a fully connected layer on top of the image embeddings, i.e., $z_i = \mW_i^T \phi(x) + b_i$, where $\mW_i$ and $b_i$ are weights and bias for target label, respectively. Let $q(k|x)$ denote the ground-truth distribution over classes for this training example such that $\sum_{i=k}^{\gK} q(k|x) = 1$. As one image may have multiple ground-truth labels, $q(k|x)$ is uniformly distributed to the ground-truth labels. 
The cross-entropy loss for the example is computed as: 
\begin{equation}
\ell = -\sum_{k=1}^{\gK} log(p(k|x))q(k|x)
\end{equation}

While softmax loss works well when the number of classes is not large (say 10K or 100K), several challenges arise if the number of classes is increased to millions or even billions. First, the computational complexity involved in computing the normalization constant of the target class probability $p(k|x)$ is prohibitively expensive~\cite{bengio2008adaptive,jean2014using}. Second, as the training objective encourages the logits corresponding to the ground-truth labels to be larger than all other logits, the model may learn to assign full probability to the ground-truth labels for each training example. This would result in over-fitting and make the model fail to generalize~\cite{szegedy2016rethinking}. 

% sampled softmax
Instead of computing the normalization constant for all the classes, we sample a small subset of the target vocabulary $L \in \{1, \dots, \gK\}$ to compute the normalization constant in $p(k|x)$ for each parameter update. Then, the target label probability can be computed as:
\begin{equation}
p'(k|x) = \frac{exp\{z_k\}}{\sum_{i \in L} exp\{z_i\}}
\label{eq:sampled_softmax}
\end{equation}
which leads to much lower computational complexity and allows us to efficiently use Tensor Processing Units (TPUs)~\cite{jouppi2017datacenter} to train this deep image embedding model with sampled softmax.

% label smoothing
Furthermore, to discourage the model from assigning full probability to the ground-truth labels (and therefore becoming prone to over-fitting), we follow~\cite{szegedy2016rethinking} to ``smooth'' the label distribution by replacing the label distribution with a mixture of the original ground-truth distribution $q(k|x)$ and the fixed distribution $u(k)$: 
\begin{equation}
q'(k|x) = (1 - \eps)q(k|x) + \eps u(k)
\end{equation}\label{eq:label_smoothing}
where $u(k)$ is a uniform distribution over the sampled vocabulary $u(k)=\frac{1}{|L|}$, and $\eps$ is a smoothing parameter.

Finally, the discriminative objective for training the neural network can be defined as the cross-entropy of the target label probability on the sampled subset and the smoothed ground-truth distribution: 
\begin{equation}
    \gL(\vtheta) = -\sum_{x_{i} \in \gD_L} \sum_{k \in L_i} log(p'(k|x_i))q'(k|x_i)
    \label{eq:supervised_loss}
\end{equation}
where $\vtheta$ denotes the neural network parameters. The ground-truth labels of $x_i$ are always selected within the sampled labels $L_i$. In our experiments, we randomly sample 100K classes for each training instance and $\eps$ is selected to be $0.1$.

\subsection{Neural Graph Learning}\label{subsec:Neural Graph Learning}
\begin{figure*}[t]
\centering
      \includegraphics[scale=0.4,bb= 0 0 1133 433]{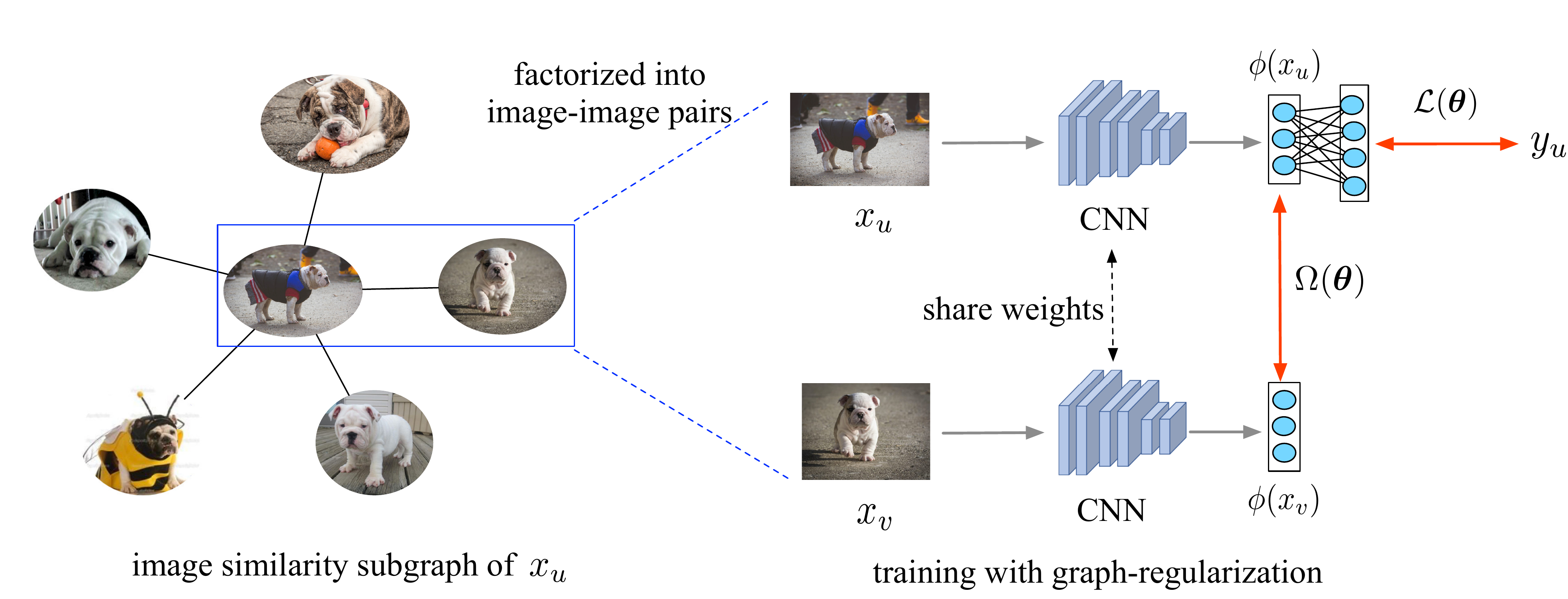}
      \vspace{-6pt}
  \caption{An illustration of a graph-regularized neural network. The image similarity subgraph of a training image $x_u$ (with the ground-truth labels $y_u$) is factorized into image-image pairs, where the neighbor image $x_v$ is semantically similar to $x_u$. The training objective consists of both the supervised loss $\mathcal{L}$ and the graph regularization $\Omega$; minimizing $\Omega$ drives the distance between the embeddings of similar images---$\phi(x_u)$ and $\phi(x_v)$---to be minimized, which means the neural network is trained to encode the local structure of a graph.}
 \label{fig:ngm_image_eg}
\end{figure*}

% \textbf{Graph-regularized classification loss}
% \reminder{dacheng pause here: complete NGM after graph-construction}
% We argue that the discriminative objective in~\cite{eq:supervised_loss} only takes into account the images that have been returned $\mathcal{D}_V$, while the huge amounts of visual signals and the relational structure of noisy labels in the noisy image corpus $\mathcal{D}$ are not fully exploited. 
While the discriminative objective indeed paves the way to learning an image representation that captures fine-grained semantics, there is more information available in human interactions with images. Many of these additional data sources can be represented as graphs (such as image-image co-occurrence), and yet current vision models (\eg, ResNet) cannot consume such graphs as inputs. Thus, we propose to train the network using graph structure about the relationships among images. In particular, images that are more strongly connected in the graph should reflect stronger semantic similarity based on user feedback (see Section~\ref{subsec:Graph Construction} for the details of graph construction), and should be closer in the embedding space. To achieve this goal, we deploy graph regularization~\cite{bui2018neural} to train the neural network for encouraging neighboring images (from the graph) to lie closer in the embedding space. 
The final graph-regularized objective is the sum of the discriminative loss and the graph regularization:
\begin{equation}
    \gR(\vtheta) = \gL(\vtheta) + \alpha \underbrace{\sum_{(u,v)\in \gE} w_{u, v} d\Big(\phi(x_u), \phi(x_v)\Big)}_{\Omega(\vtheta)}
    \label{eq:total loss}
\end{equation}
\noindent where $\Omega(\vtheta)$ denotes the graph regularizer, $\gE$ represents a set of edges between images, $w_{u, v}$ represents the edge weight between image $u$ and $v$, $\phi(\cdot)$ is the representation extracted from the embedding layer\footnote{In general, the embedding layer refers to the layer right before the softmax layer.}, $d(\cdot)$ is the distance metric function, and $\alpha \geq 0$ is the multiplier (applied on regularization) that controls the trade-off between the discriminative loss and the regularization induced from the graph structure. An illustration of the graph-regularized neural network is given in Figure~\ref{fig:ngm_image_eg}.

The multiplier $\alpha$ (applied on regularization) controls the balance between the discriminative information (\ie, predictive power) and the contributions of the graph structure (\ie, encoding power). In other words, the neural network is trained to both (a) make accurate classification (or prediction), and (b) encode the graph structure. When $\alpha = 0$, the proposed objective ignores the graph regularization and degenerates to a neural network with only supervised objective in Eq.~(\ref{eq:supervised_loss}). On the other hand, when $p'(x) = \phi(x)$, where $p'(x)$ is the predicted label distribution, we have a label propagation objective as in~\cite{ravi2016large} by training with the objective using $p'(x)$ directly without parameters $\vtheta$ (\ie, no neural network involved), and letting the distance function $d(\cdot)$ and the loss function $\ell(\cdot)$ to be mean squared errors (MSE). The label propagation objective encourages the learned label distribution $p'(x)$ of similar images to be close. Furthermore, the label distribution of a sample is aggregated from its neighbors to adjust its own distribution. Thus, the proposed objective in Eq.~(\ref{eq:total loss}) could be viewed as a ``graph-regularized version of neural network objective'' or as a ``non-linear version of label propagation.''

\subsection{Graph Construction}
\label{subsec:Graph Construction}
In this section, we provide the details for constructing graphs used by the regularizer $\Omega(\vtheta)$ in Eq.~(\ref{eq:total loss}). In addition to image-query pairs (described in Section~\ref{sec:Problem Formulation}) where images are annotated by queries for obtaining semantics, we propose to find ``image-image'' pairs where the semantics shared by these two images are closer to human perception and beyond textual search queries. Each image is treated as a ``vertex'' and an image-image pair is treated as an ``edge,'' together forming a graph that can be included as ancillary training data. 

% Co-click graph
Specifically, each image-image pair contains one source vertex $x_u \in \gD_L$ and one target vertex $x_u \in \{\gD_L \bigcup \gD_U \}$. In this work, we introduce two methods to construct edges: (a) based on co-click rate of the image pair, and (b) based on similar-image click rate of the image pair. The co-click rate of the image pair characterizes how often users select both the source image $x_u$ and the target image $x_v$ in response to both $x_u$ and $x_v$ being concurrently identified by search results from a textual search query. This type of image-image relationship sheds light on the question: ``Given that one image is selected from  the resulting images, what other images that are sufficiently similar will also be selected?'' If the co-click rate between $x_u$ and $x_v$ is higher than a pre-defined threshold, $x_u$ and $x_v$ are considered to be sufficiently similar and an edge between them is constructed; the edge weight $w_{u, v}$ is calculated based on the co-click rate. Then $x_u$ and $x_v$ will be used for calculating the graph regularization $\Omega(\vtheta)$ in Eq.~(\ref{eq:total loss}).

% VSI graph
Different from the co-click rate, the similar-image click rate of the image pair characterizes how often users select the source image $x_u$ in response to $x_u$ being identified by a search result for a search query using the target image $x_v$ (instead of a textual query). This type of image-image relationship sheds the light upon the question: ``Given an image issued as the query, what other images that are sufficiently similar will be selected in response to the query image?'' Similar with how edges are constructed based on the co-click rate, if the similar-image click between $x_u$ and $x_v$ is higher than a pre-defined threshold, $x_u$ and $x_v$ are considered to be sufficiently similar and an edge between them is constructed. Edge weight $w_{u, v}$ is calculated based on the similar-image click rate.

\section{Training Framework}
\label{sec:architecture}
\begin{figure*}[tbp]
  \includegraphics[scale=0.45,bb=0 0 982 500]{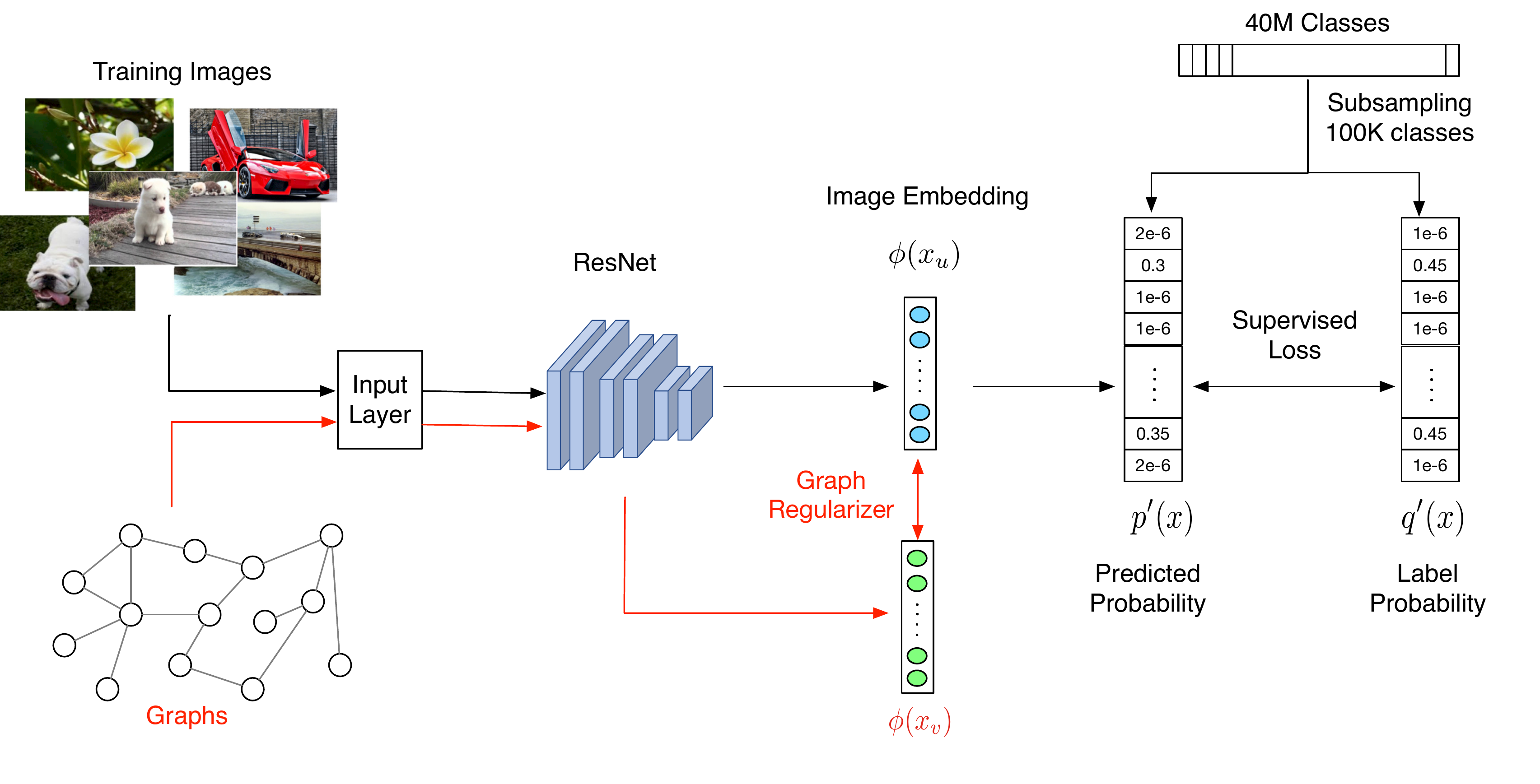}
  \vspace{-16pt}
  \caption{An illustration of the Graph-RISE framework. Flow in red is added to enable graph regularization and required only during training. In the input layer, a labeled image is associated with one of its neighbor images, which can be either labeled or unlabeled, and then fed into the ResNet together with its neighbor image. Then, the image embeddings generated from ResNet are used to both (a) compute the cross-entropy loss and (b) graph regularization. 
  }
  \label{fig:graph_regularized_resnet}
\end{figure*}

In this section, we provide the details of network architecture and training infrastructure along with the configurations we used in this work.
\subsection{Network Architecture}
\label{subsec:Network Architecture}

Figure~\ref{fig:graph_regularized_resnet} illustrates the proposed network architecture. The main model is the 101-layer ResNet (referred as ResNet-101) \cite{he2016deep}. Compared to Inception \cite{szegedy2015going}, ResNet-101 has larger model capacity, which yields more than 2\% of performance improvement on our internal metric for embedding evaluation. While the major architecture of ResNet-101 remains unchanged, several detailed configurations have been modified. The input layer is modified to take enlarged input images from 224$\times$224 to 289$\times$289 pixels. The output 10$\times$10$\times$2K feature map is first avg pooled to 4$\times$4$\times$2K using a 4$\times$4 kernel of stride 2, and then flattened and projected to 64-dimensional layer representing image embeddings. The activation function is ReLU-6\footnote{ReLU-6 computes Rectified Linear Unit as: $\min(\max(x, 0), 6)$.}. Finally, a softmax layer is added to produce a multinomial distribution across 40 million classes (\eg, queries).

\begin{comment}
First, the input layer is modified to take enlarged input images from 224$\times$224 to 289$\times$289 pixels. Next, the size of average pooling is modified from (3, 3) square to (4, 2) rectangle. The last average pooling layer is then flattened and connected to a 64-dimensional layer representing image embeddings. The activation function is ReLU-6\footnote{ReLU-6 computes Rectified Linear Unit as: $\min(\max(x, 0), 6)$.}. Finally, a softmax layer is added to produce a multinomial distribution across 40 million classes (\eg, queries). 
\reminder{Modify the description: "ResNet 101 with 289x289 input. The output 10x10 x 2k feature map is first avg pooled to 4x4x2k using a 4x4 kernel of stride 2, and then flattened and projected to 64-dim embedding."}
\end{comment}

During training, both training samples and their neighbors provided from graphs (described in Section~\ref{subsec:Graph Construction}) are fed into the model for enabling graph regularization; the 64-dimensional embedding layer is selected as the target layer to be regularized as described in Eq.~(\ref{eq:total loss}). %Furthermore, since calculating a large softmax (40 million) is computationally challenging, sampled softmax ~\cite{bengio2008adaptive}---an important sampling technique for handling large output spaces---is applied during training. 
Furthermore, 100K out of 40 millions labels are sampled via an important sampling technique~\cite{bengio2008adaptive} for each parameter update Eq.~(\ref{eq:sampled_softmax}). Finally, batch normalization is applied during training.

During the inference phase, a 64-dimensional L2 normalized embedding~\footnote{L2 normalization is not applied on embedding during training as it makes the training hard to converge.} is generated for each input image as a new, semantically-meaningful representation. Note that neighbors and graph regularization is not required when making inference (see the flow in red in Figure~\ref{fig:graph_regularized_resnet}). In addition to the embedding, the queries with the top-$k$ predicted probabilities are also outputted. Since the focus of this paper is image embedding, the output queries will largely be ignored in the rest of this paper.

\subsection{Training Infrastructure}
\label{subsec:Training Infrastructure}
We implement the network architecture described in Section~\ref{subsec:Network Architecture} using TensorFlow\cite{abadi2016tensorflow}. The details of training configurations are as follows. We select the batch size to be 24, and the momentum \cite{qian1999momentum} as the optimizer; the initial learning rate is 0.001 and will be decayed with an exponential rate of 0.9 every 100,000 steps. The label smoothing $\epsilon$ is 0.1. The multiplier for applying L2 regularization (a.k.a. ``weight decay'') is 0.00004. For configurations related to graph regularization, the multiplier for applying graph regularization $\alpha$ is 1.0, and the distance function $d(\cdot)$ in Eq.~(\ref{eq:total loss}) is selected to be cosine distance\footnote{We have also experimented with using the Euclidean (L2) distance as $d(\cdot)$; when using Euclidean distance with $\alpha=0.01$, it achieves almost the same performance as using cosine distance with $\alpha=1$}. For constructing graphs, the threshold is set to $0.1$, and we combine the edges (approximately 50 million edges, built from co-clicks and similar-image clicks) into one graph for calculating regularizer $\Omega(\vtheta)$ in Eq.~\ref{eq:total loss}.

Since our model is one of the largest vision models in terms of number of parameters (40M $\times 64$ plus the parameters of ResNet-101 architecture), together with the scale of the training dataset, using TPUs~\cite{jouppi2017datacenter} to train the model is the only feasible solution. The training is distributed to 8$\times$8 TPU cores, and takes two weeks to converge from scratch after 5M steps. Training with graph regularization costs additional computation that grows with the number of neighbors used (in this work, approximately 40\% training time increase due to the use of neighbors). To reduce the computation, we compare the performance of two models: one is trained without graph regularization from scratch; the other is trained with graph regularization with $\alpha$ set to zero until it is converged, and then set $\alpha=1$ to fine-tune the model. We find that the performance of the first model (w/o graph regularization) is slightly better in the beginning of the training, and two models achieves almost the same performance after 4M steps. With these setting, the fine-tuning model (the one with graph regularization) takes approximately 2 additional days to train for 500K extra steps until convergence.

\section{Experiments}
\label{sec:experiment}
In this section, we explain the details of evaluation setup, and then show the experimental results for performance evaluation. We also provide the case studies for the qualitative analysis.
% \reminder{Futang helps finish Experimental setting}
\subsection{Evaluation Setup}
\label{subsec:Evaluation Setup}
We are interested in providing image embeddings with instance-level semantics, such that the similarity between embeddings approximates the relationships (of images) in terms of semantics. To evaluate the performance of the proposed image embedding model, we conduct both k-Nearest-Neighbor (kNN) search and triplet evaluation as metrics; these two are the most popular methods used for evaluating embedding models. 

For kNN evaluation, we conduct experiments on ImageNet~\cite{krizhevsky2012imagenet} and iNaturalist~\cite{van2018inaturalist} datasets, and then report two metrics in the experiments: Top-1 and Top-5 accuracy, where Top-k accuracy calculates the percentage of query images that find at least 1 image---from the top k searched image results---carrying the exact same labels as the query images. For the ImageNet dataset, the images in the validation set are used as the query images, and the training set is used as the index set to search for top-k results. For the iNaturalist dataset, for each class two images are randomly sampled to construct the query set, and the remainder of the images in that class are used as the index set.
% \reminder{Describe how we split the dataset: what's the query set adn what's the index set.}

For triplet evaluation, we follow the evaluation strategy in ~\cite{wang2014learning} to sample triplets $(A, P, N)$---representing Anchor, Positive, Negative images---from Google Image Search and ask human raters to verify if P is more semantically closer to A than N. We sample the triplets in a way such that A and P have the same or a very similar instance-level concept, and N is slightly less relevant (``hard-negative''). Each triplet is independently ranked by three raters and only the triplets receiving unanimous scores are used for evaluation. Assume positive image P is rated to be more similar to anchor image A than negative image N, the prediction of a model is considered to be accurate if the following condition holds: 
\begin{equation}
    \eta + d(\phi(A), \phi(P)) - d(\phi(A), \phi(N)) < 0
    \label{eq:margin_eta}
\end{equation}
\noindent where $\eta$ is the hyper-parameter that controls the margin between the distance of two image projections.

Two triplet datasets are created for calculating triplet evaluation metrics: (a) Product-Instance-Triplets (PIT) is a dataset designed to focus on evaluating the semantic concepts of images in the commercial product vertical, which consists of 10,000 triplets; (b) Generic-Instance-Triplets (GIT) is a dataset focusing on evaluating the semantic concepts of general images, including all possible image verticals from Google Image Search, consisting of 14,000 triplets.

\subsection{Model Comparisons}
\label{subsec:Comparison Methods}

We compare the proposed method with the following state-of-the-art models: 
\begin{itemize}
    \item DeepRanking model~\cite{wang2014learning} that employs triplet loss on multi-scale Inception network architecture~\cite{szegedy2016rethinking} with an online triplet sampling algorithm. 
    \item Inception network architecture~\cite{szegedy2016rethinking} that employs sampled softmax loss over 8 millions labels.
    \item ResNet-101 network architecture~\cite{he2016deep} that employs sampled softmax loss over 8 millions and 40 millions labels (referred as ResNet (8M) and ResNet (40M) in Table~\ref{tab:exp_knn}, respectively). 
    \item Graph-RISE model based on ResNet-101 network architecture proposed in Section~\ref{subsec:Neural Graph Learning}. 
\end{itemize}
The input layers in DeepRanking model and Inception model both use 224$\times$224 image pixels, while the ResNet-101 and Graph-RISE use 289$\times$289. Label smoothing is applied to all the classification-based models. When the graph regularization multiplier $\alpha=0$, Graph-RISE is equivalent to the ResNet-101 model. In all the experiments, the Euclidean (L2) distance of the embedding vectors extracted from the penultimate layer---the layer before the final softmax or ranking layer---is used as similarity measure. To evaluate the effectiveness of image embeddings, no individual fine-tuning is performed for each dataset, and all the experiments are conducted directly based on the learned embeddings of the input images.

% \reminder{Describe the architecture for comparison methods.}
% \reminder{Compare the performance of using co-click graph, VSI graph, and merged graph.}

% \begin{itemize}
%     \item{KNN evals: ImageNet (testing data), iNaturalist (publicly available)}
%     \item{Triplet evals: internal human-rated (e.g., PIT) triplets for evaluation.}
% \end{itemize}
% % \subsection{Evaluation Metrics}
% % Remove this subsection if space not permitted.
% \begin{itemize}
%     \item{Accuracy: e.g., how to calculate KNN/triplet accuracy.}
%     \item{Precision & Recall}
%     \item{Show comparison w/ Wang Jiang's paper \cite{wang2014learning}}
% \end{itemize}

\subsection{Performance Evaluation}
\begin{table}[t]
\begin{tabular}{l|ll|ll}
                 & \multicolumn{2}{c}{ImageNet~\cite{krizhevsky2012imagenet}}  & \multicolumn{2}{c}{iNaturalist~\cite{van2018inaturalist}}      \\
                 & \multicolumn{1}{c}{Top-1} & \multicolumn{1}{c}{Top-5} & \multicolumn{1}{c}{Top-1} & \multicolumn{1}{c}{Top-5} \\ \hline
DeepRanking     & 35.20                      & 60.93                     & 6.03                      & 13.71                     \\
Inception (8M)   & 61.92                      & 84.77                   & 17.30                     & 34.58                     \\
ResNet (8M)      & 62.49                     & 84.65                     & 17.36                     & 34.15                     \\
ResNet (40M)     & 66.20                      & 86.41                     & 27.05                     & 47.44                     \\
Graph-RISE (40M)     & 68.29                     & 87.75                     & 31.12                     & 52.76                    
\end{tabular}
\caption{Performance comparisons (in \%) via kNN search accuracy on publicly available datasets.}
\label{tab:exp_knn}
\end{table}

\begin{table}[tbp]
\begin{tabular}{l|ll}
                 & PIT  &  GIT     \\ \hline
DeepRanking & 74.95              &  77.25            \\        
Inception (8M)   & 82.81                  & 86.54                       \\
ResNet (8M)      & 85.46                     & 87.72                     \\
ResNet (40M)     & 86.70                      & 88.90                    \\
Graph-RISE (40M)     & 87.16                & 89.53                    \\
\end{tabular}
\caption{Performance comparisons (in \%) via triplet accuracy ($\eta=0$) on the internal evaluation datasets.}
\label{tab:exp_triplet}
\end{table}

\begin{figure}[tbp]
  \includegraphics[width=0.9\linewidth,bb=0 0 1006 612]{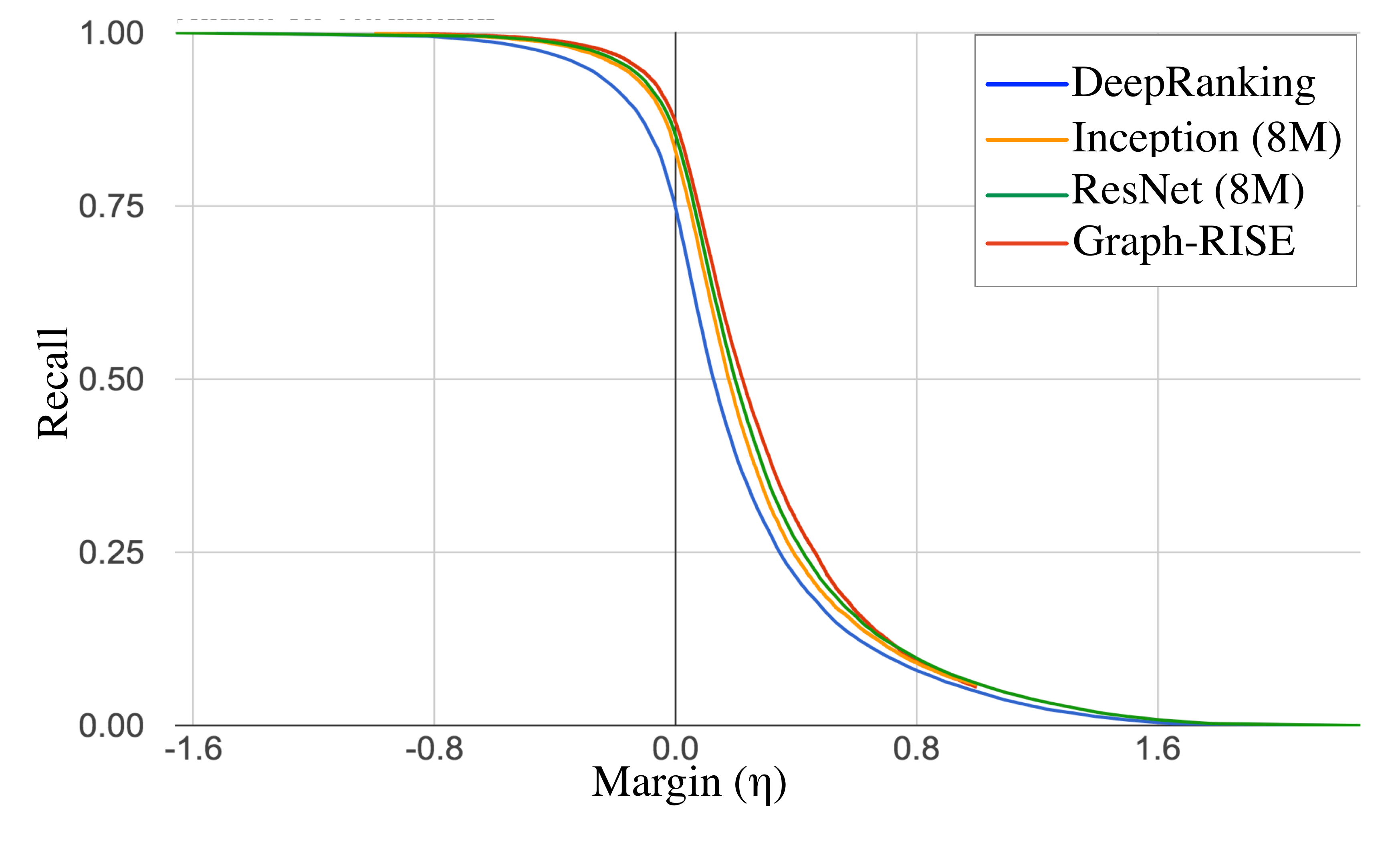}
  \caption{PIT triplet evaluation on Recall v.s. Margin.}
  \label{fig:wtb_pr}
\end{figure}

\begin{figure}[tbp]
  \includegraphics[width=0.9\linewidth,bb=0 0 1004 621]{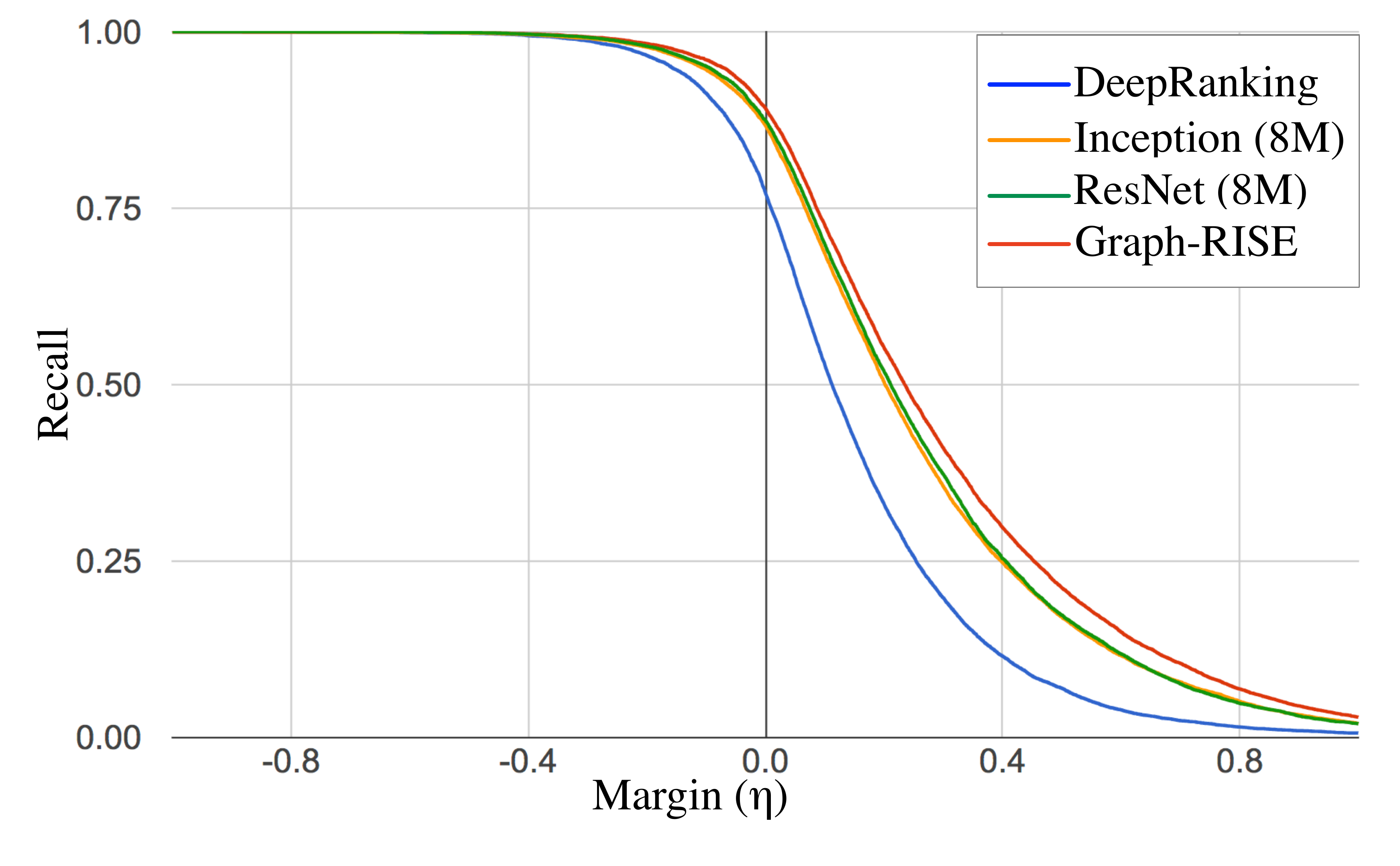}
  \caption{GIT triplet evaluation on Recall v.s. Margin.}
  \label{fig:vsi_pr}
\end{figure}

\begin{figure*}[tbp]
  \includegraphics[scale=1,bb=0 0 501 395]{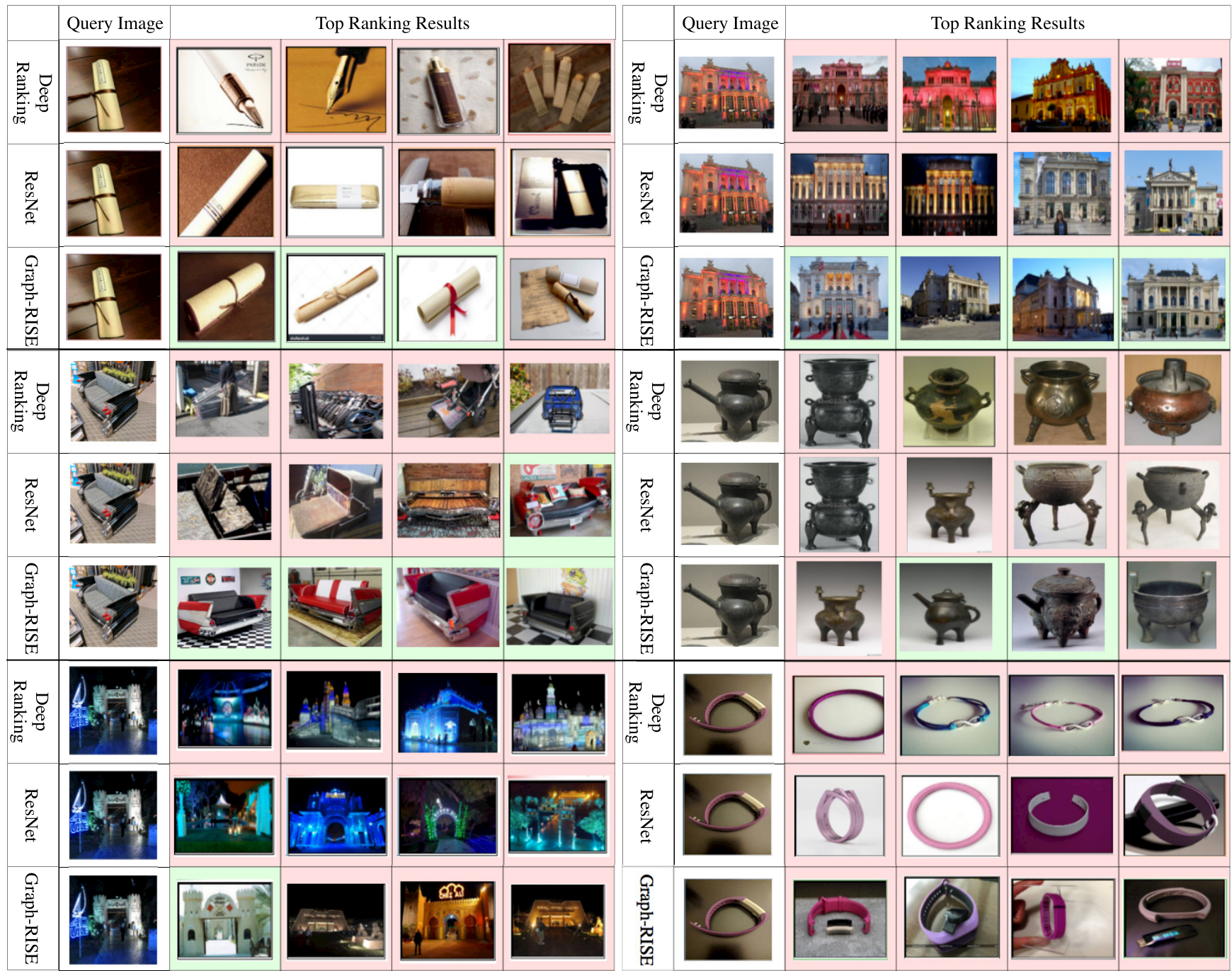}
  \caption{Retrieval results for 6 randomly-chosen query images. For each query image, we provide the Top-4 images retrieved (from 1 billion images) by DeepRanking, ResNet (40M), and Graph-RISE. Each retrieval result is rated and color-coded by human raters: retrieved images colored by green are rated to be strongly similar with the query image, whereas images colored by red are rated to be not (or somewhat) similar. Notice that images retrieved by Graph-RISE generally conform to experts' ratings.}
  \label{fig:exp_examples}
\end{figure*}

Table~\ref{tab:exp_knn} provides the performance comparisons (in terms of percentage) on kNN evaluations, and Table~\ref{tab:exp_triplet} shows the triplet evaluations (also in terms of percentage). From these results, we have several observations. First, {\ourmodel} significantly outperforms the previous state-of-the-art \cite{wang2014learning} and other models without graph regularization in all the evaluation criteria. We attribute this to the fact that {\ourmodel} leverages the graph structure via neural graph learning to drive the embeddings of similar images to be as close as possible. Notice that, compared to the previous state-of-the-art \cite{wang2014learning}, {\ourmodel} {\bf improves the Top-1 accuracy by almost 2X} (from 35.2\% to 68.29\%) on ImageNet dataset, and by {\bf more than 5X} (from 6.03\% to 31.12\%) on the iNaturalist dataset. 

Second, compared to the Inception network architecture, training image embeddings with ResNet-101 improves the performance for most datasets. This confirms the observation from \cite{sun2017revisiting} that to fully exploit $\gO(300M)$ images, a higher capacity model is required to achieve better performance. Furthermore, we confirm that sampled softmax is an effective technique to train image embedding model with datasets that have extremely-large label space; by comparing Inception (8M) and DeepRanking \cite{wang2014learning} (both based on Inception network), we observe that sampled softmax helps achieve better performance in triplet loss, even if DeepRanking directly aims at optimizing the triplet accuracy.

Moreover, increasing the number of labels (from 8M to 40M) significantly improves the kNN accuracy. We conjecture that the labels in 40M are more fine-grained than 8M, and therefore the learned image embeddings also need to capture fine-grained semantics in order to distinguish these 40M labels. In addition, we find that training ResNet-101 using larger input size (289$\times$289) instead of smaller input size (224$\times$224) also helps improve the model performance (from 85.13\% to 86.7\%, in terms of accuracy on PIT triplet evaluation), since larger input size encapsulates more detailed information from training images.

Figure~\ref{fig:wtb_pr} and Figure~\ref{fig:vsi_pr} depict the comparisons of triplet evaluation among four models: DeepRanking, Inception (8M), ResNet (8M) and {\ourmodel}, on PIT dataset and GIT dataset respectively. Note that ResNet (40M) is ignored in the figures since the curves of ResNet (40M) and {\ourmodel} are visually difficult to distinguish. In these two figures, x-axis is ``Margin'' and y-axis is ``Racall'' rate (the higher the better). ``Margin'' is the $\eta$ in Eq.~\ref{eq:margin_eta} representing the margin between the distance of ``Anchor-Negative pair'' and the distance of ``Anchor-Positive pair.'' A large margin means ``the Negative image is  further away from the Anchor image than the Positive image.'' ``Recall'' rate represents the percentage of triplets that satisfy $\eta + d(\phi(A), \phi(P)) < d(\phi(A), \phi(N))$. From these two figures, the performance of {\ourmodel} is consistently better the other models.

\subsection{Qualitative Analysis}
\label{subsec:Qualitatively Studies}
% \reminder{Da-Cheng: pauses here and adds more analysis later..}
% Provide several cherry-picked, ``wow'' examples.

Next, we evaluate the quality of images retrieved by DeepRanking \cite{wang2014learning}, ResNet (40M), and {\ourmodel} models. Given a randomly-selected query image, each method retrieves the most semantically similar images from an index containing one billion images. The top-ranked results are sent out to be rated by human experts. Figure~\ref{fig:exp_examples} illustrates the retrieval results for 6 randomly-selected query images; for each query image we provide the Top-4 images retrieved. The images colored by green are rated to be strongly similar with the query image, whereas the images colored by red  are rated to be not (or somewhat) similar. Compared to other models, images retrieved by {\ourmodel} generally conform to experts' ratings, meaning that {\ourmodel} captures the semantic meaning of images more effectively as judged by human raters. For example, given a query image of ``scroll with ribbon'' (top-left in Figure~\ref{fig:exp_examples}), the top three  images retrieved by {\ourmodel} are also ``scroll with ribbon'' (with similar colors, textures and shapes), and are rated as strongly similar by human experts. Another example is a query image of a landmark (top-right in Figure~\ref{fig:exp_examples}); {\ourmodel} is able to retrieve images of the exact same landmark, while the other methods are only able to retrieve images of somewhat similar buildings. 

In addition, we observe that the images retrieved by DeepRanking tend to be only visually similar to the query images, rather than semantically similar. This is probably because generating triplets that reflect the semantic concepts is very difficult, especially when the classes are ultra-fine-grained.

\section{Conclusion}
\label{sec:conclusion}
% This is the conclusion.

In this work, we present {\ourmodel} to answer the motivational question: \textit{``Is it possible to learn image content descriptors (a.k.a., embeddings) that capture image semantics and similarity close to human perception?''} {\ourmodel} confirms that ultra-fine-grained, instance-level semantics can be captured by image embeddings extracted from training a sophisticated image classification model with large-scale data: $\gO(40M)$ classes and $\gO(260M)$ images. {\ourmodel} is also the first image embedding model based on neural graph learning that leverages graph structures of similar images to capture semantics close to human image perception. We conduct extensive experiments on several evaluation tasks based on both kNN search and triplet ranking, and experimental results confirm that {\ourmodel} consistently and significantly outperforms the state-of-the-art methods. Qualitative analysis of image retrieval tasks also demonstrates that {\ourmodel} effectively captures instance-level semantics.

\section*{Acknowledgement}
We would like to thank Dr. Sergey Ioffe and Dr. Thomas Leung for the reviews and suggestions. We also thank Expander, Image Understanding and several related teams for the technical support.

\balance
\bibliographystyle{ACM-Reference-Format}
\bibliography{acmart}

\end{document}